# Pattern Analysis with Layered Self-Organizing Maps

**By David Friedlander**

## 1. Abstract

Abstract— This paper defines a new learning architecture, Layered Self-Organizing Maps (LSOMs), that uses the SOM and supervised-SOM learning algorithms. The architecture is validated with the MNIST database of hand-written digit images. LSOMs are similar to convolutional neural nets (covnets) in the way they sample data, but different in the way they represent features and learn. LSOMs analyze (or generate) image patches with maps of exemplars determined by the SOM learning algorithm rather than feature maps from filter-banks learned via backprop.

LSOMs provide an alternative to features derived from covnets. Multi-layer LSOMs are trained bottom-up, without the use of backprop and therefore may be of interest as a model of the visual cortex. The results show organization at multiple levels. The algorithm appears to be resource efficient in learning, classifying and generating images. Although LSOMs can be used for classification, their validation accuracy for these exploratory runs was below the state of the art. The goal of this article is to define the architecture and display the structures resulting from its application to the MNIST images.

## 2. Background

The mathematics of the SOM training algorithm [Kohonen 01] is not completely understood [Polani 01]. It has been shown, however, that SOMs are not equivalent to training networks with a reasonably smooth cost function [Heskes 99]. For this study, the training algorithm was

implemented as

$$\bar{\phi}_{i,j}(t+1) = \bar{\phi}_{i,j}(t) + L_r(t)N(r,c,i,j)\left[\bar{\zeta} - \bar{\phi}_{i,j}(t)\right] \tag{1}$$

where $\bar{\phi}_{i,j}(t)$ is the SOM vector at grid position $i, j$, and time $t$, where $\bar{\phi}_{i,j}(0)$ is a vector with small, random components.

The *learning rate* is $L_r(t) = 0.9\exp\left(\dfrac{-t}{t_m}\right)$, where $t_m$ is the number of iterations, and the

*neighborhood function* is $N(r,c,i,j) = \exp\left(\dfrac{-\hat{d}(r,c,i,j)}{R(t)^2}\right)$, where

$$r, c = index \max_{i,j} d\left(\overline{\zeta} - \overline{\phi}_{i,j}(t)\right)^2 \qquad (2)$$

$d(\cdot, \cdot)$ is the Euclidean distance.

The random variable $\overline{\zeta} \in \left\{\overline{\zeta}_i\right\}$ is a randomly selected input vector. The indices $r$ and $c$ represent the row and column of the closest SOM vector match. The function $R(t) = 0.5s \exp\left(\dfrac{-t \log(0.5s)}{t_m}\right)$, where $s$ is the number of points per side of the SOM grid. The function

$$\hat{d}(r_1, c_1, r_2, c_2) = \begin{cases} \sqrt{(r_1 - r_2)^2 - (c_1 - c_2)^2}, \text{if} < R(t)^2 \\ \qquad\qquad 0 \text{ otherwise} \end{cases}.$$

Single layer SOMs have been used for classification using the *supervised SOM* algorithm [Kohonen 01]. During training, additional dimensions were added to the input vectors, one per class. The components were set to zero except for the one corresponding to the correct class. This causes the SOM vectors to cluster around class labels.

## 3. Theory

Large classes of data can be represented as vector fields over rectangular lattices. This includes video, for example, which can be represented on a three-dimensional lattice containing two spatial and one temporal dimension [Goodfellow 2016]. Black and white video would be represented as a scalar field over the lattice and color video as a 3-dimensional vector field.

If a SOM is trained on random samples of a high-dimensional time series, it can then be used to create a corresponding low (typically 2) dimensional time series by replacing each input feature, in order, with a low dimensional vector corresponding to the matched SOM grid point. The dynamics of the low-dimensional time series reflects the dynamics of the high-dimensional inputs.

The inputs to the next higher level are the coordinates of matches on the SOM grid from the level below. These high-dimensional vectors are matched against the corresponding SOM to create a 2-dimensional time series as input to the next higher level. The SOM training algorithm organizes the SOM vectors over the SOM grid points so that nearby paths over the grid tend to be similar in terms of the set of SOM vectors the paths connect.

The above concept is generalized to inputs that are vector or scalar fields over square lattices. This is done by creating the *scan* operator, which is a generalization of the dimension-raising algorithm described above. This operator covers the input lattice with a set of (possibly overlapping) regions and creates a high-dimensional vector for each region, which is the outer product of all the components of all the low-dimensional data in that region. This results in a smaller lattice of high dimensional vectors. The SOM *match* operator is then applied to the high-dimensional lattice by matching each high-dimensional vector with the SOM grid to find the closest match. A two-dimensional vector corresponding to the row and column of the match replaces the high-dimensional vectors on the input lattice.

Each layer of the architecture contains both operators, allowing for the definition of a recursive architecture with a potentially arbitrary number of layers. The architecture is trained one layer at a time, bottom up. The input to the bottom layer is the data and the input to all other layers is the ordered set of match vectors from the layer below. For object identification, the top layer is designed so there is only one feature vector per object. *Supervisory channels* are added to the top feature during training to represent the class labels. They cause the top-level SOM vectors to cluster around exemplars of the classes. After training, the supervisory components are used to label each top-level SOM node with a class label, the index of the maximum component. The class labels are used to facilitate clustering rather than for optimization of a cost function. This seems to make the technique less susceptible to over-training.

## 3.1   The Scan operator

The scan operator transforms vector fields over lattices. The result is a vector field of higher dimension over a smaller lattice, in terms of points per side, but of the same dimensionality. First the scan operator divides the lattice into potentially overlapping regions which span it. It then assigns a high dimensional vector to each region. The high-dimensional vector in a given region is the outer product of all the components of all the initial vectors in that region. The match operator substitutes a 2-dimensional vector, corresponding the SOM grid match, for each high-dimensional on the input lattice. The levels are trained one-at-a-time starting at the bottom. The inputs to higher levels correspond to fixed length paths through the grid of the level below.

It is defined as follows: $S : \overline{x}_{ij} \rightarrow \overline{\phi}_{mn}$, where $\overline{x}_{ij}$ is the vector at row *i*, column *j*, of the initial lattice, and $\overline{\phi}_{mn}$ is the vector of row *m* column *n* of the transformed lattice. Here two indices are shown for the input lattice, but the number of actual indices will equal the dimension of the input lattice. Two will be shown for the rest of this document.

Figure 1 shows examples of the scan operator scanning one and two-dimensional lattices. The operation is defined as follows: let *s* be the number of points per side in the initial grid, *p* be the number of points per side in the scanning window, and *v* be the *stride*, the number of points the window is shifted for each sample. Then *u*, the number of points per side in the transformed lattice is $u = \dfrac{s - p}{v} + 1$, and $\dim(\phi) = p^n \dim(x)$, where *n* is the dimension of the input lattice. The parameter *u* is must be an integer, which puts restrictions on *p* and *v* given *s*.

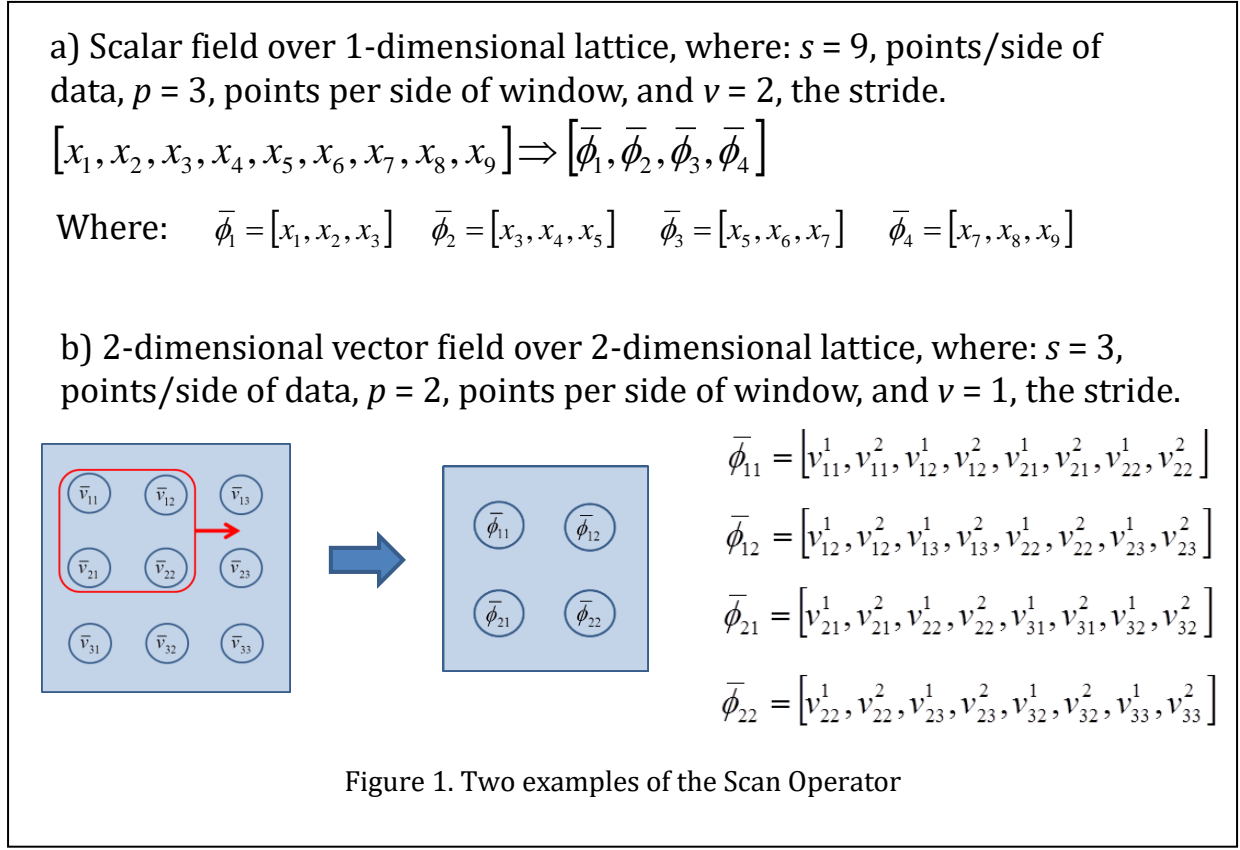

a) Scalar field over 1-dimensional lattice, where: $s = 9$, points/side of data, $p = 3$, points per side of window, and $v = 2$, the stride.

$$\left[ x_1, x_2, x_3, x_4, x_5, x_6, x_7, x_8, x_9 \right] \Rightarrow \left[ \overline{\phi}_1, \overline{\phi}_2, \overline{\phi}_3, \overline{\phi}_4 \right]$$

Where: $\quad \overline{\phi}_1 = \left[ x_1, x_2, x_3 \right] \quad \overline{\phi}_2 = \left[ x_3, x_4, x_5 \right] \quad \overline{\phi}_3 = \left[ x_5, x_6, x_7 \right] \quad \overline{\phi}_4 = \left[ x_7, x_8, x_9 \right]$

b) 2-dimensional vector field over 2-dimensional lattice, where: $s = 3$, points/side of data, $p = 2$, points per side of window, and $v = 1$, the stride.

$$\overline{\phi}_{11} = \left[ v_{11}^1, v_{11}^2, v_{12}^1, v_{12}^2, v_{21}^1, v_{21}^2, v_{22}^1, v_{22}^2 \right]$$

$$\overline{\phi}_{12} = \left[ v_{12}^1, v_{12}^2, v_{13}^1, v_{13}^2, v_{22}^1, v_{22}^2, v_{23}^1, v_{23}^2 \right]$$

$$\overline{\phi}_{21} = \left[ v_{21}^1, v_{21}^2, v_{22}^1, v_{22}^2, v_{31}^1, v_{31}^2, v_{32}^1, v_{32}^2 \right]$$

$$\overline{\phi}_{22} = \left[ v_{22}^1, v_{22}^2, v_{23}^1, v_{23}^2, v_{32}^1, v_{32}^2, v_{33}^1, v_{33}^2 \right]$$

Figure 1. Two examples of the Scan Operator

The process is similar to scanning in convolutional networks [LeCun 89] except, instead of resulting in a set of feature maps, the scan operator takes the outer product (a high-dimensional vector) of the scan window and, after training, derives a single map (the SOM gird) of high-dimensional *patterns* (SOM vectors).

## 3.2　The SOM Match operator

The match operator is defined as $M\left(\overline{\phi}, \overline{\Sigma}_{ij}\right) = r\hat{x}_1 + c\hat{x}_2$, where $r, c = index \max_{i,j} d\left(\overline{\phi} - \overline{\Sigma}_{i,j}\right)^2$ i.e., it finds the *closest match* as was done in training. $\overline{\Sigma}_{ij}$ is the SOM grid, $\dim\left(\overline{\phi}\right) = \dim\left(\overline{\Sigma}\right)$ and $d\left(\cdot, \cdot\right)$ is the Euclidian distance function. When applied to a lattice, $M : \overline{\phi}_{ij} \rightarrow \overline{x}_{mn}$, where $\dim\left(x\right) = 2$. Unlike the scan operator, it does not change the number of points per side of the transformed lattice.

## 3.3　The Layered SOM Architecture

Each layer consists of a *scan* operator $S$, followed by a *match* operator $M$. A *layer* index is now added in the upper right-hand side of the quantities that depend on layer. For example, $\overline{x}_{ij}^m$ is the input to layer $m$. The architecture is then defined by $\overline{x}_{ij}^{m+1} = M\left(S\left(\overline{x}_{ij}^m\right)\right)$, where $\overline{x}_{ij}^0$ is the input to layer 0, i.e., the input data. Also, at the top level satisfies $\overline{x}^{L-1}$ is a single vector, not a lattice of vectors.

## 3.4    Training

The architecture is trained from bottom up, one level at a time. The scanned data are input to the bottom level. The output of each level is a 2-dimensional vector field of two-dimensional matches over the lattice of the scanned input. The output of one level is the input to the level above until the top level. The architectures in this article are designed for object recognition. There are many features per input at the bottom level and one feature per input (a 1x1 lattice) at the top level. The lattice size shrinks at each level until it contains only one element at the top level. The top level is trained using the class labels as per the *supervised SOM* algorithm.

### 3.4.1    Bottom Level

Each input is initially represented by a vector (or scalar) field over a lattice. After the scan operation, each input is now represented by $S : \bar{x}_{ij}^0 \to \bar{\phi}_{mn}^0$, where the indices $i$ and $j$ range over the input lattice (for a 2-d lattice), and the indices $m$ and $j$ over the sub-window regions. The unordered (i.e. randomly chosen) set of all $\phi^0$ vectors from all inputs is represented by $\left\{ \bar{\phi}_{mn}^0 \right\} = \left\{ S\left( \bar{x}_{ij}^0 \right) \right\}$. This set of feature vectors is used to train $\Sigma_0$, the SOM at layer zero. After training, the *ordered* output of layer 0 is the input to layer 1: $\bar{x}_{mn}^1 = M\left( \bar{\phi}_{mn}^0 \right)$. See Figure 2.

### 3.4.2    Middle levels

The output each level is the input to the level above. These inputs consist of lattices of 2d vectors with decreasing numbers of points per side at each level. Thus the SOM at level $0 < i < L$, where $\Sigma^i$ is trained on $\left\{ \bar{x}_{mn}^i \right\} = \left\{ M\left( S\left( \bar{x}_{ij}^{i-1} \right) \right) \right\}$. After training, the output $\bar{x}_{mn}^{i+1} = M\left( S\left( \bar{x}_{ij}^i \right) \right)$ is produced from the ordered input. See Figure 3.

### 3.4.3    Top level

In order to define the training operation, an *input-number* index is added to the input feature at the top level of the SOM. Since there is only one feature per input at the top level, the lower indices corresponding to the input lattice position are dropped and replaced with the input index, for the rest of this section.

At the top level, there is one feature channel per class: $\phi_h \to \tilde{\phi} = \phi_h \otimes \bar{p}(h)$, where $p_i = \delta\left( i, label(h) \right)$. Once training is complete, the supervisory channels of the SOM grid are not used in matching unlabeled inputs, but they are used to assign a class label to each node on the top level grid, the index of the maximum component of the supervisory channels: $Class(x, y) = index \max_i p_i(r, c)$. The number of classes must not be less than the number of nodes in the top level SOM. Typically, there would be multiple nodes per class and therefore multiple simulated inputs, per class. See Figure 4.

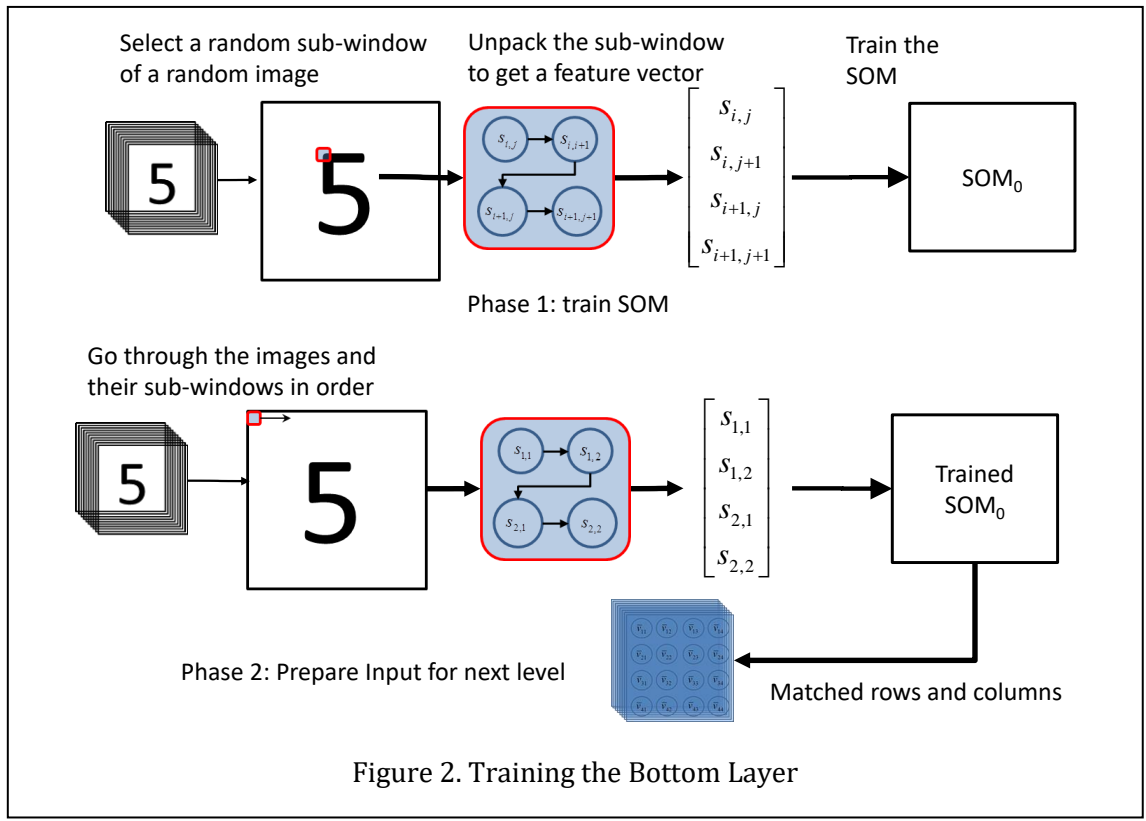

Figure 2. Training the Bottom Layer

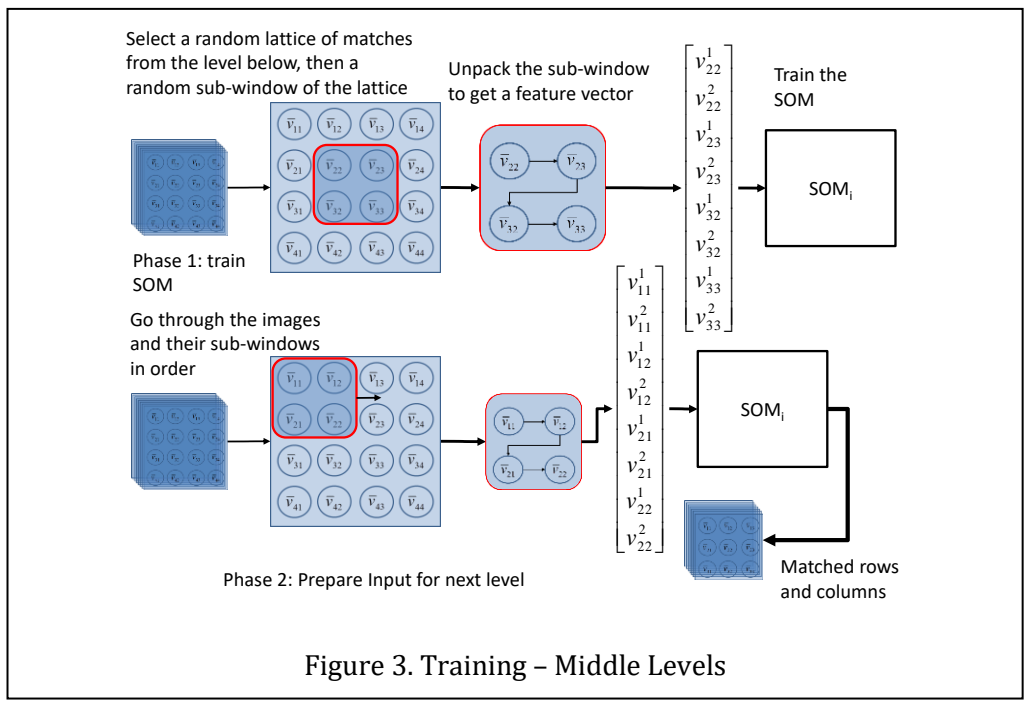

Figure 3. Training – Middle Levels

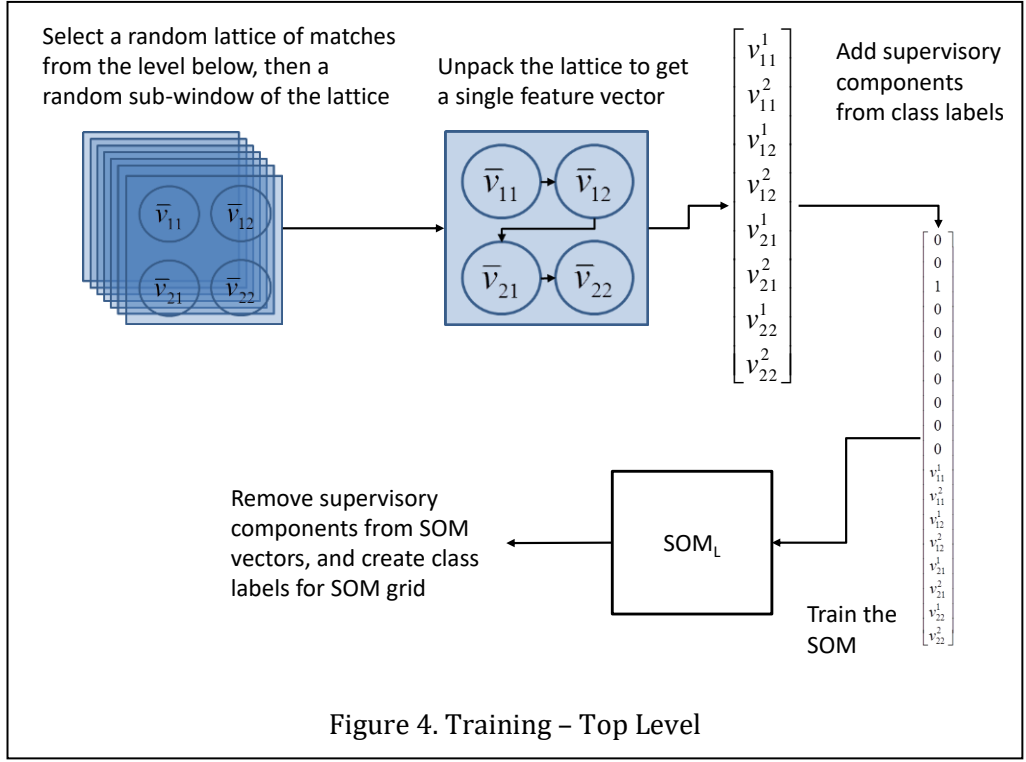

Figure 4. Training – Top Level

## 3.5    Data Generation

When an input is matched to the architecture, the scan operator converts it to many low-level (but high-dimensional) features, after matching the SOM at level 0; the input becomes two-dimensional features distributed over a lattice with many points. At each level, the number points per side of the input lattice is decreased by the scan operator to create high-dimensional input to the SOM match operator, until there is only one feature per input at the top level, which matches a single grid point of the SOM at the top level, where each grid point has a class label.

This process can be approximately reversed resulting in a simulated input for each top-level grid point. The approximate match inverse is $\tilde{M}\left(\overline{y}, \overline{\Sigma}\right) = \overline{\Sigma}_{y_1 y_2}$, where $y_1, y_2$ are the components of $\overline{y}$ corresponding to a valid grid location in $\overline{\Sigma}$. The inverse match operator retrieves the SOM vector associated with the pair of indices. The vector $\overline{\psi} \equiv \tilde{M}\left(\overline{y}, \overline{\Sigma}\right)$ can be rearranged to represent the a window with $p$ points per side where $\dim(\psi) = p^n \dim(y)$, where $n$ is the dimension of the input lattice (and the window). The $\psi$ vectors are the simulated equivalent of the $\phi$ vectors that are produced during classification. For the lower levels, inputs and outputs are lattices of low dimensional vectors. Since the input to level $h$ is the output of level $h$-1, the high-dimensional vectors at level $h$, $\overline{\psi}_{mn}^{h}$, can be unpacked to provide simulated input vectors $\overline{y}_{ij}^{h-1}$ for the level below.

The approximate scan inverse for lattices is more complex than a simple unpacking for two reasons. First, the match inverse requires integer coefficients because they are used to find SOM vectors at the level below. The output of the inverse scan operator should therefore also have integer components but the SOM vectors have real coefficients. This is due to the averaging of the SOM training algorithm. Second, the windows can overlap. When they are unpacked, there can be multiple, different values for the components of the simulated input lattice. The inverse scan operator $\tilde{S} : \bar{\psi}_{mn}^h \rightarrow \bar{y}_{ij}^{h-1}$, unpacks each $\psi$ vector into a sub-window, collects all the values for each input vector on the input lattice. It then averages all multiple values and rounds each component of each input vector. Finally, the recursive simulation architecture can be expressed as

$\bar{y}_{ij}^{h-1} = \tilde{S}\left(\tilde{M}\left(\bar{y}_{kl}^h\right)\right)$, where $\bar{y}_{ij}^0$ is the simulated input data and $\bar{y}_{ij}^{L-1}$ is the initial top-level grid point (and associated class label) used to initiate the simulation.

## 4. Application

This section applies the theory to image classification, in particular, the MNIST data [LeCun, et.al. 2015]. Each architecture is defined as an ordered list of one or more *layers*, and each layer is defined by three parameters: *s*, number of nodes per side of the input lattice, *p* the number of nodes per side of the scanning window, *v*, the stride, and *k*, the number of nodes per side of the SOM grid. The specification for an *n-layer* architecture would take the form

$$((p_0, v_0, k_0), (p_1, v_1, k_1), \ldots (p_{n-1}, v_{n-1}, k_{n-1})).$$

This section shows the results for a number of LSOM networks. The SOM vectors at each level are displayed as two dimensional images over the corresponding two dimensional SOM grid. The generated data are 28x28 images over the SOM gird at the top level and finally, the class labels are also displayed over the top level grid.

### 4.1    Single-layer systems

A single-layer system is equivalent to a traditional *supervised SOM* [Kohonen 2001]. The input data for each image are organized a scalar field over a 28 x 28 node lattice. Table 1 shows the training and validation accuracies, and the run times of a single layer systems as a function of number of iterations, number of images used for training, and the number of nodes per side in the SOM grid. Accuracy goes up with all three input parameters, but most strongly with the size of the SOM grid. On average, validation accuracy is only slightly less than training accuracy. Figure 5 shows validation accuracy as a function of SOM grid size and number of images for 50,000 training iterations. Figures 6 and 7 show the class labels and generated images.

Table 1.  Single Layer Runs

| Iterations | Images | Side | Train | Validate | Tim |
|---|---|---|---|---|---|
| 1000 | 1000 | 20 | 0.808 | 0.725 | 16.34 |
| 1000 | 1000 | 50 | 0.894 | 0.829 | 43.10 |
| 1000 | 1000 | 100 | 0.896 | 0.804 | 02:18. |
| 1000 | 10000 | 20 | 0.7706 | 0.7538 | 02:11. |
| 1000 | 10000 | 50 | 0.8427 | 0.8283 | 02:43. |
| 1000 | 10000 | 100 | 0.8584 | 0.8361 | 05:02. |
| 1000 | 20000 | 20 | 0.77125 | 0.7569 | 04:09. |
| 1000 | 20000 | 50 | 0.8373 | 0.82535 | 05:01. |
| 1000 | 20000 | 100 | 0.8408 | 0.83115 | 08:13. |
| 10000 | 1000 | 20 | 0.884 | 0.795 | 25.57 |
| 10000 | 1000 | 50 | 0.972 | 0.87 | 01:08. |
| 10000 | 1000 | 100 | 0.996 | 0.881 | 03:47. |
| 10000 | 10000 | 20 | 0.8344 | 0.8154 | 02:11. |
| 10000 | 10000 | 50 | 0.9116 | 0.8974 | 03:07. |
| 10000 | 10000 | 100 | 0.9353 | 0.915 | 06:39. |
| 10000 | 20000 | 20 | 0.8337 | 0.82375 | 04:15. |
| 10000 | 20000 | 50 | 0.9049 | 0.8968 | 05:27. |
| 10000 | 20000 | 100 | 0.9249 | 0.91235 | 09:55. |
| 50000 | 1000 | 20 | 0.874 | 0.793 | 34.62 |
| 50000 | 1000 | 50 | 0.996 | 0.868 | 02:35. |
| 50000 | 1000 | 100 | 1 | 0.862 | 09:36. |
| 50000 | 10000 | 20 | 0.8355 | 0.8255 | 02:24. |
| 50000 | 10000 | 50 | 0.9266 | 0.9027 | 04:40. |
| 50000 | 10000 | 100 | 0.9637 | 0.9357 | 12:34. |
| 50000 | 20000 | 20 | 0.84055 | 0.8318 | 04:33. |
| 50000 | 20000 | 50 | 0.91965 | 0.9098 | 07:07. |
| 50000 | 20000 | 100 | 0.9559 | 0.9368 | 15:48. |

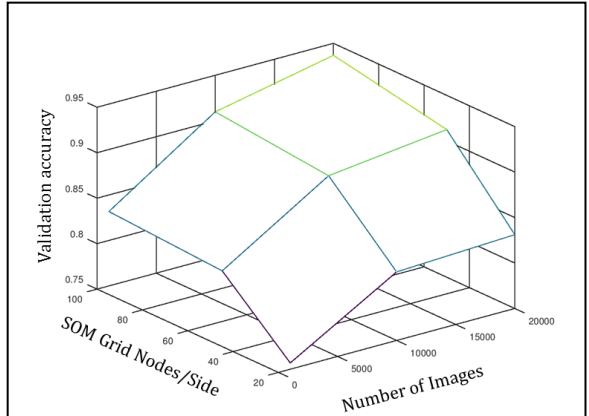

Figure 5.  Single Layer Accuracy

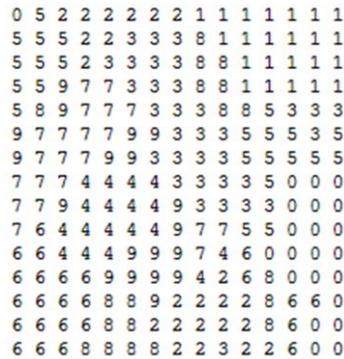

Figure 6. Single Layer Class Labels for a 15x15 SOM grid

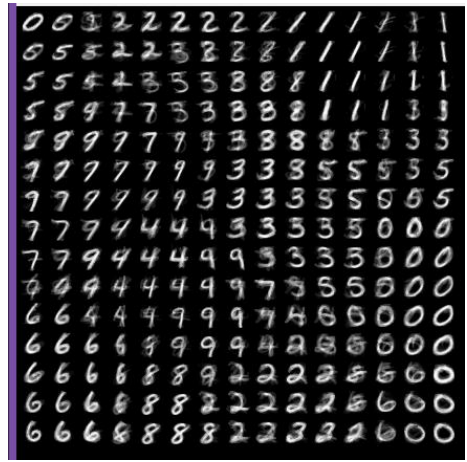

Figure 7. Single Layer Generated Images for a 15x15 SOM Grid

## 4.2   Two-layer systems

For two-layer systems, in the first layer, the data are scanned by a square sub-image window defined by the length of the window side, and the *stride*, or shift between successive sub-images.

The number of sub-images per image for layer 0 is: $n_0 = \left(\dfrac{28 - p_0}{v_0} + 1\right)^2$ .  In the second layer,

$p_1 = n_0$, $n_1 = 1$ and the stride is irrelevant since $p_1 - n_0 = 0$ .  Typically, small windows and a stride

of 1 produce better results.  Table 2 shows results for all valid architectures with sub-windows of up to 7 pixels per side using 1000 images, 50,000 iterations, and 50x50 SOM grids in both layers.

Figures 8 through 11 show detailed results for a 7x7 sub-window with a stride of 1 using 50,000 iterations and 50,000 images.  The bottom SOM grid is 50x50, but the top SOM grid is 15x15, so all of the simulated images can fit in a figure.  Figure 8 and 9 show the class labels and reconstructed images.  Figure 10 shows the bottom-level SOM vectors corresponding to scalar features over a 7x7 window.  The second layer contains SOM vectors that correspond to 2-dimensional vector features over a 22x22 window.  In order to display a 2-dimensional field, RGB colors were used with the green component set to 0, the red component set to the first component of the vector and the blue component set to the second component.  The results are shown in Figures 8 through 11.  The information in level 0 contains the primitive features and level 1 contains a (smaller) red/blue image of the generated digits.  This happened on all LSOMs tested, although the exact color scheme varies depending on the network architecture.  The reason is not known.

### Table 2.  Two-Layer Runs

| iterations | images | Train | Validate | sub-side1 | stride1 | grid-side1 | sub-side2 | grid-side2 |
|---|---|---|---|---|---|---|---|---|
| 50000 | 1000 | 0.9279 | 0.9109 | 2 | 1 | 50 | 14 | 50 |
| 50000 | 1000 | 0.8445 | 0.8244 | 2 | 2 | 50 | 14 | 50 |
| 50000 | 1000 | 0.8712 | 0.8372 | 3 | 1 | 50 | 26 | 50 |
| 50000 | 1000 | 0.9179 | 0.8972 | 4 | 1 | 50 | 25 | 50 |
| 50000 | 1000 | 0.8874 | 0.8596 | 4 | 2 | 50 | 13 | 50 |
| 50000 | 1000 | 0.8744 | 0.847 | 4 | 3 | 50 | 9 | 50 |
| 50000 | 1000 | 0.7106 | 0.6827 | 4 | 4 | 50 | 7 | 50 |
| 50000 | 1000 | 0.9191 | 0.9013 | 5 | 1 | 50 | 24 | 50 |
| 50000 | 1000 | 0.9228 | 0.8999 | 6 | 1 | 50 | 23 | 50 |
| 50000 | 1000 | 0.9052 | 0.8877 | 6 | 2 | 50 | 12 | 50 |
| 50000 | 1000 | 0.9252 | 0.9131 | 7 | 1 | 50 | 22 | 50 |
| 50000 | 1000 | 0.8737 | 0.8427 | 7 | 3 | 50 | 8 | 50 |
| 50000 | 1000 | 0.6454 | 0.6137 | 7 | 7 | 50 | 4 | 50 |
| 50000 | 50000 | 0.79714 | 0.8054 | 7 | 1 | 50 | 22 | 15 |
| 50000 | 50000 | 0.76416 | 0.7642 | 7 | 1 | 10 | 22 | 20 |
| 50000 | 50000 | 0.62252 | 0.6268 | 7 | 1 | 20 | 22 | 10 |

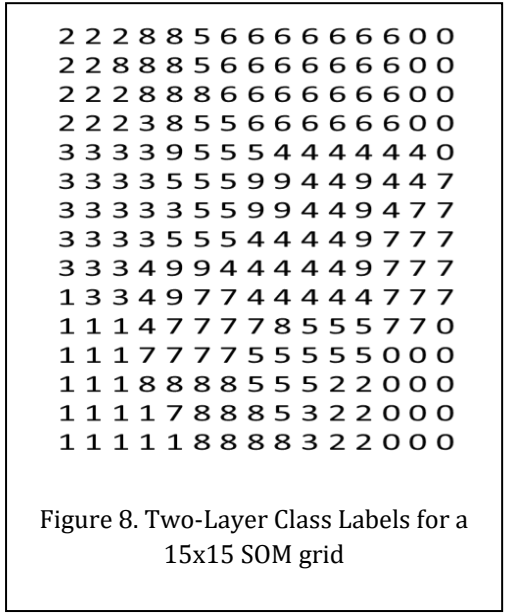

Figure 8. Two-Layer Class Labels for a
15x15 SOM grid

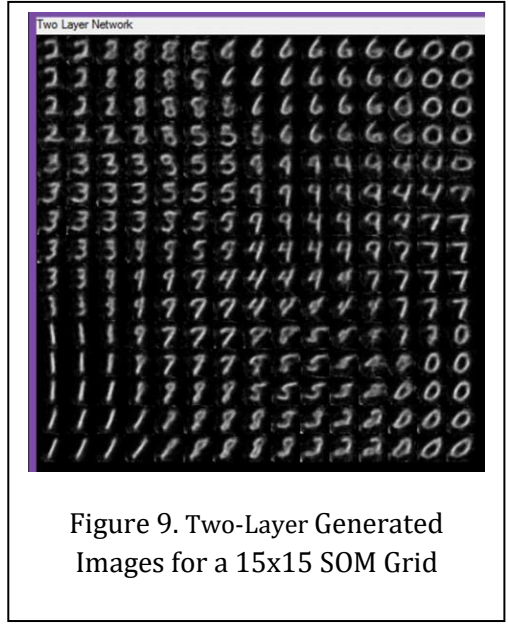

Figure 9. Two-Layer Generated
Images for a 15x15 SOM Grid

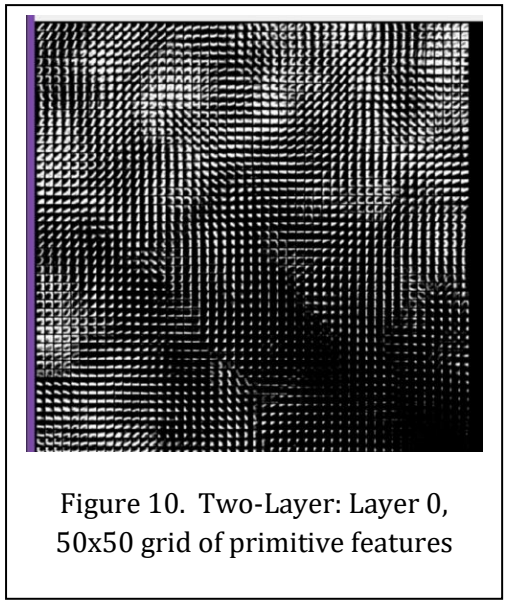

Figure 10.  Two-Layer: Layer 0,
50x50 grid of primitive features

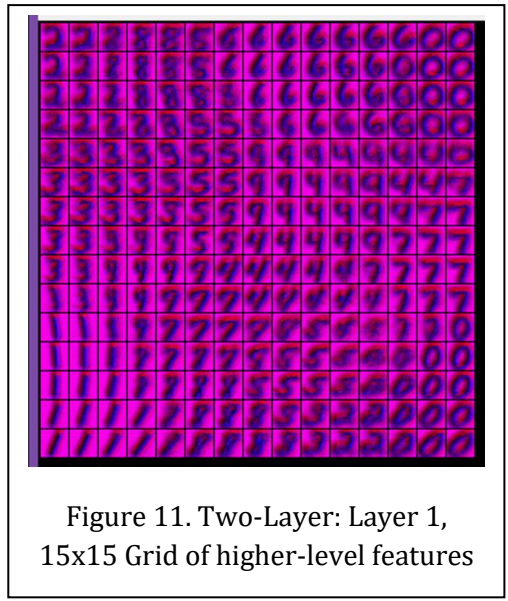

Figure 11. Two-Layer: Layer 1,
15x15 Grid of higher-level features

## 4.3    Greater than two layer systems

The first layer of the architecture is unique because its input is the data stream and the top layer is
unique because there is only one feature vector per image and its SOM grid has class labels.  The
middle layers are all similar.  They each take in a two-dimensional lattice of two-dimensional
vectors: the row and column indices of the corresponding SOM grid matches and produce a smaller
two dimensional lattice of two-dimensional vectors.  Table 3 shows the results for a set of 3-layer
architectures.

Detailed results are shown for a single four level run whose architecture described in Table 4.
Figure 12 shows the 4x4 pixel primitive features on level 0's 45x45 grid and Figures 13 through 19
show the features and projections for each of the other three levels.

### Table 3.  Three-Layer Runs

|            |        |         |          |           |        | Layer 0   |           |        | Layer 1   |           | Layer 2   |
| iterations | images | Train   | Validate | data-side | stride | grid-side | data-side | stride | grid-side | data-side | grid-side |
|------------|--------|---------|----------|-----------|--------|-----------|-----------|--------|-----------|-----------|-----------|
| 50000      | 1000   | 0.8662  | 0.8312   | 7         | 1      | 50        | 7         | 3      | 50        | 6         | 50        |
| 50000      | 1000   | 0.8815  | 0.8465   | 7         | 1      | 50        | 7         | 1      | 50        | 16        | 50        |
| 50000      | 1000   | 0.9077  | 0.8773   | 4         | 1      | 50        | 7         | 1      | 50        | 19        | 50        |
| 50000      | 50000  | 0.87412 | 0.8596   | 7         | 1      | 50        | 7         | 3      | 50        | 6         | 100       |
| 50000      | 50000  | 0.62504 | 0.6266   | 7         | 1      | 10        | 7         | 3      | 10        | 6         | 10        |
| 50000      | 50000  | 0.65314 | 0.657    | 7         | 1      | 10        | 7         | 3      | 10        | 6         | 10        |
| 50000      | 50000  | 0.80084 | 0.8014   | 7         | 1      | 10        | 7         | 3      | 20        | 6         | 30        |

### Table 4.  Four-Layer Run

| Train | Valid. | Images | Iterate | Sample | Stride | SOM | Layer |
|-------|--------|--------|---------|--------|--------|-----|-------|
| 0.905 | 0.875  | 10000  | 50000   | 4      | 1      | 45  | 0     |
|       |        |        |         | 4      | 1      | 50  | 1     |
|       |        |        |         | 4      | 2      | 55  | 2     |
|       |        |        |         | 10     | 10     | 60  | 3     |

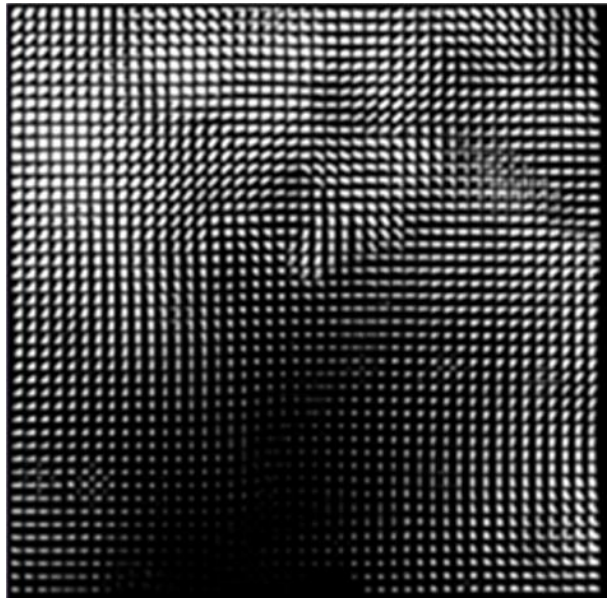

Fig 12. 4x4 Primitive Features on a 45x45 Grid

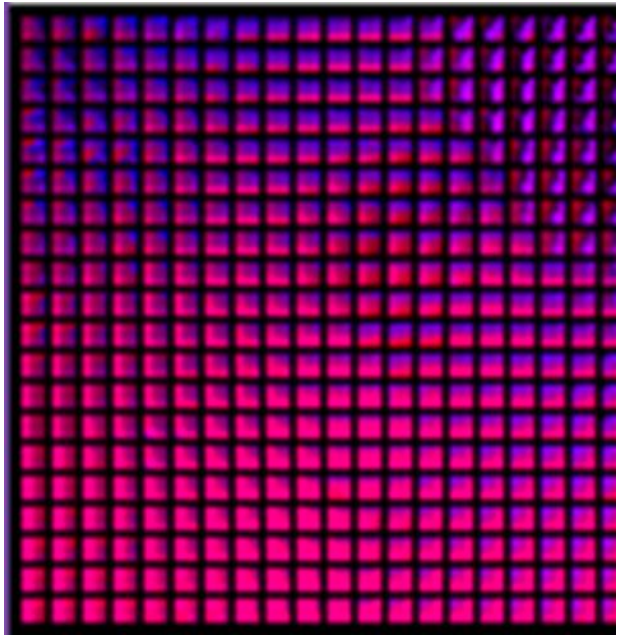

Fig 13 Level 1 Features for Rows & Columns 15 – 34

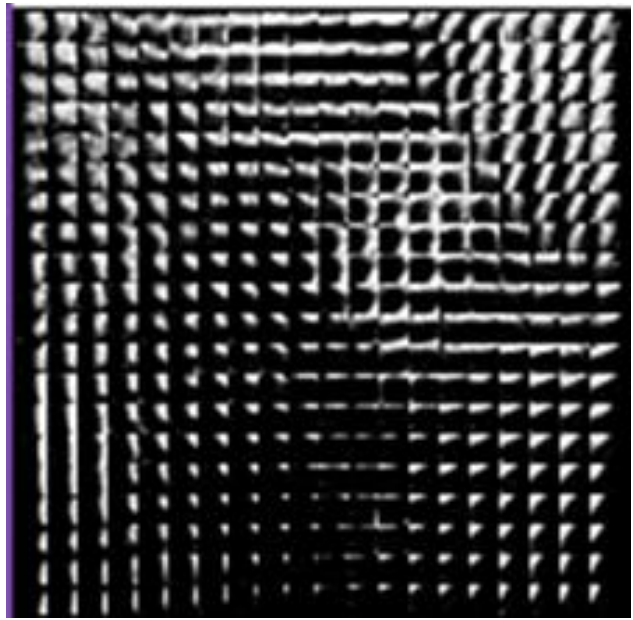

Fig 14. Level 1 Projections, Rows & Columns 15-34.

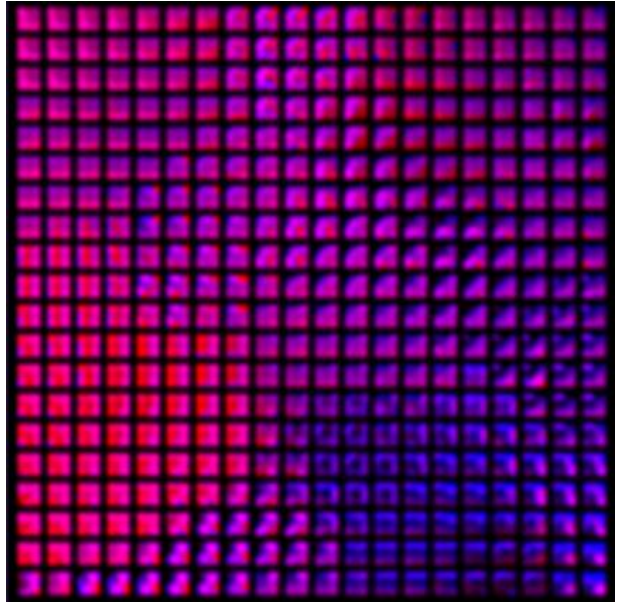

Fig 15. Level 2 Features for Rows & Columns 15 – 34

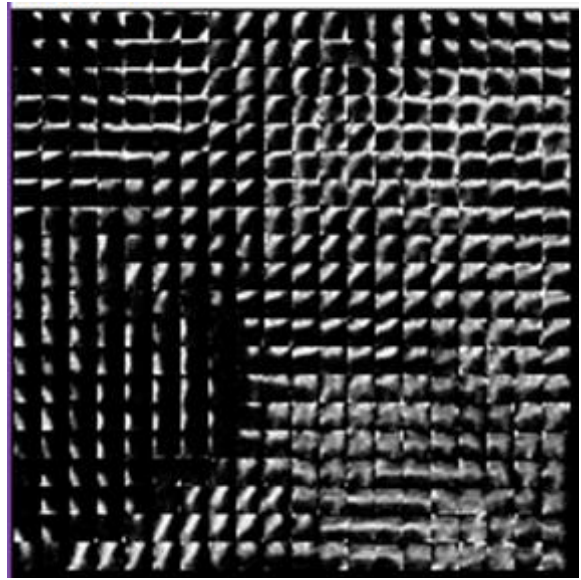

Fig 16. Level 2 Projections: Rows & Columns 15-34

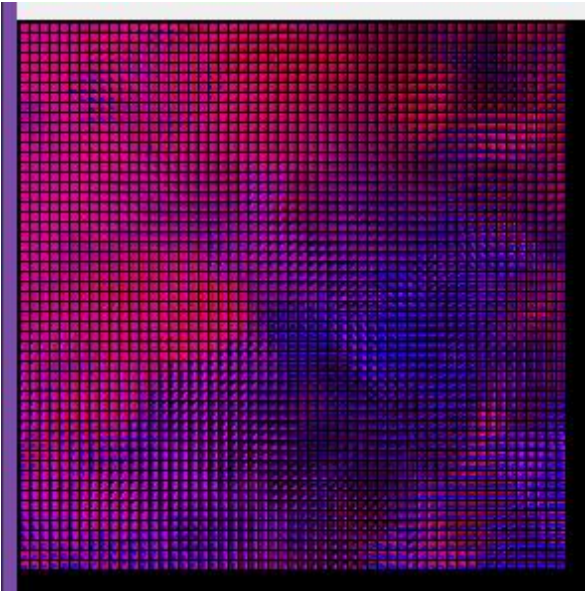

Fig 17. Level 2 Features for the Entire Grid

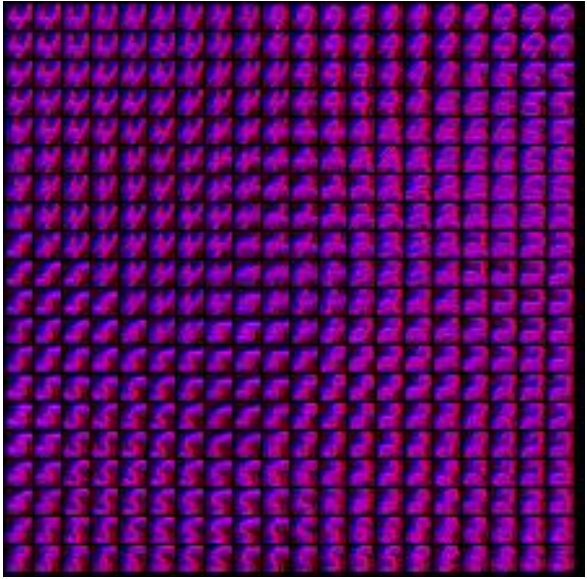

Fig 18.  Top Level Features, Rows Columns 15-34

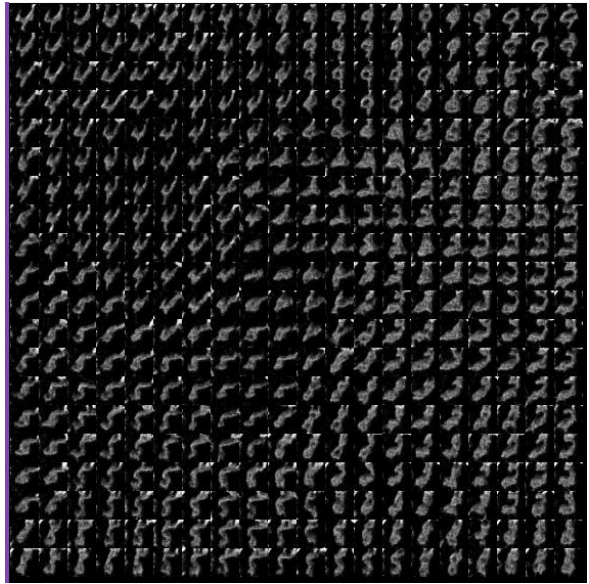

Fig 19. Top Level Projections, Rows & Columns 15-34

Figs. 18 and 19 show the close-up for the top level grid and the top level projections. The patches in Fig. 18 contain 10x10 abstracted versions of the 28x28 digits possibly analogous to the lowest frequency component of wavelet decomposition [Kohonen 2001].

## 5. Conclusions and Future Work

The mathematics of the SOM training algorithm [Kohonen 2001] is not completely understood [dPolani 2001]. There is evidence, however, that SOMs are not equivalent to training networks with a smooth cost function [Heskes 1999]. This suggests that the features derived by LSOMs may be significantly different than those from covnets. LSOMS:

- use *exemplars* to create a multi-scale decomposition of the image, rather than *filter banks,*
- transmit the address space (SOM locations) of the matching exemplars up to the next level,
- allow bottom-up training for multi-level networks.
- create arrays of features at increasing scales, similar to convolutional networks
- produce an approximation of their inputs, similar to reverse convolutional networks.

The architecture has relatively little over-training without external regularization by using clustering rather than optimization over an energy function. The implementations provide control of the scaling and quantization over each level. The architecture is recursive and stores knowledge/features that have semantic interpretations and a visual form. Features appear to be organized at multiple scales within each grid as shown in the SOM grid displays.

In order to provide a semantic interpretation of the information stored in the SOM grids, let $M(\bullet) \rightarrow location(pattern(\bullet))$ and $S(\bullet) \rightarrow sequence(sample(\bullet))$. That is, the *Match* and *Scan* operators are divided into two parts each. The *pattern* finds the closest SOM vector and *location* finds the location of that vector on the SOM grid. Similarly, *sample* takes a window of data and *sequence* takes each element of window and created a single vector. Given those definitions, the semantic interpretation is:

$$location(pattern(sequence(sample(... location(pattern(sequence(sample(data))))))))$$

LSOMs have not achieved high accuracy on the MNIST data. Attempts to improve performance by adding layers were unsuccessful. Architectures with large SOM grids perform better. More research and more powerful hardware may improve the results as was the cases for the deep multi-layer perceptron.

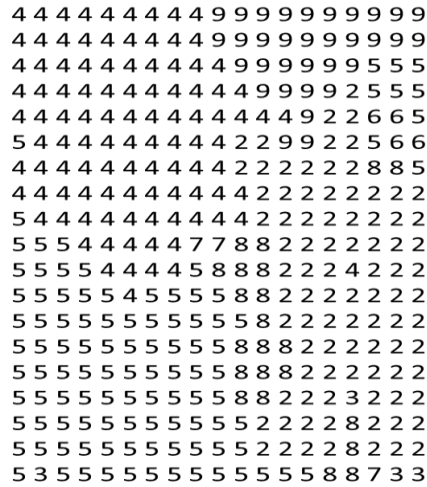

Fig. 20. Run 2: Class Map, Rows &Columns 15-33

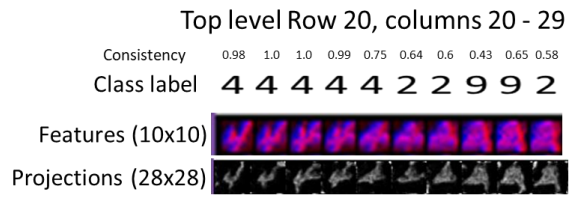

Fig. 21. Consistency vs Features and Projections

The consistency of each class label is determined by the set of labeled training images that matched a given node in the upper-level SOM. The class label is the class with the largest number of matches and the consistency is the ratio of this number to the total. The top-level SOM vector components corresponding to inconsistent nodes could be set to a values so no features will match them. The match will go to the next closest node.